# Generative 3D Cardiac Shape Modelling for In-Silico Trials

Andrei GASPAROVICI [a, b,c1] and Alex SERBAN [b,d]
[a] *Babeș-Bolyai University, Cluj-Napoca*
[b] *Advanta, Siemens SRL, Brașov*
[c] *Tiberiu Popoviciu Institute of Numerical Analysis, Cluj-Napoca*
[d] *Transilvania University of Brașov*
ORCiD ID: Andrei Gasparovici https://orcid.org/0009-0009-6150-5540

**Abstract.** We propose a deep learning method to model and generate synthetic aortic shapes based on representing shapes as the zero-level set of a neural signed distance field, conditioned by a family of trainable embedding vectors with encode the geometric features of each shape. The network is trained on a dataset of aortic root meshes reconstructed from CT images by making the neural field vanish on sampled surface points and enforcing its spatial gradient to have unit norm. Empirical results show that our model can represent aortic shapes with high fidelity. Moreover, by sampling from the learned embedding vectors, we can generate novel shapes that resemble real patient anatomies, which can be used for in-silico trials.

## 1. Introduction

In-silico trials provide an opportunity to accelerate testing of medical devices, accounting for a wider range of patients compared to in-vitro trials. To enable in-silico trials, it is necessary to have access to a pool of diverse patient data or have the ability to generate synthetic data. For example, the simulation of Transcatheter Aortic Valve Implantation procedures can be performed computationally on distinct aortic shapes representing diverse human anatomies. In this paper, we introduce a deep learning (DL) based method for generating 3D aortic valve shapes. DL has significant advantages over conventional methods such as statistical shape models (SSMs), allowing the reuse of learned representations for distinct tasks, including conditioning other generative processes, such as calcium deposit generation, and predicting hemodynamic measurements using physics-informed DL models.

Previously, Hoeijmakers et al. [3] used a SSM to reconstruct aortic geometries from CT images and to predict the pressure drop across the aortic valve by performing computational fluid dynamics simulations. Similarly, Verstraeten et al. [7] used a SSM to generate new aortic shapes, represented as deformations of a template shape. The synthetic shapes are used to generate virtual cohorts which are similar to real cohorts and considered anatomically plausible for in-silico trials.

---

[1] Corresponding Author: Andrei Gasparovici, andrei.gasparovici@ubbcluj.ro

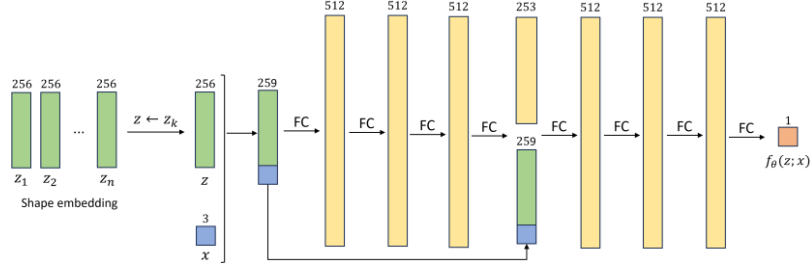

**Figure 1.** Auto-decoder neural network architecture for aortic shape modeling and generation.

A three-dimensional shape $S$ can be represented as the zero-level set of a function $f\colon \mathbb{R}^3 \to \mathbb{R}$, that is:
$$S = \{x \in \mathbb{R}^3 \colon f(x) = 0\}. \tag{1}$$
For a watertight shape (i.e., a three-dimensional surface having a well-defined interior and no holes), a suitable choice for $f$ is the signed distance function (SDF), which is a continuous mapping that assigns to each point in space the distance to the surface, with the sign indicating whether the point is inside (positive) or outside (negative) the shape (Park et al. [5]). If $f$ is known, the zero-level set can be recovered using the well-established Marching Cubes meshing algorithm (Lorensen and Cline [4]).

Using this method to represent shapes, Park et al. [5] introduced a DL-based method for shape modeling and generation by representing a family of watertight shapes $S^{(k)}$ as the zero-level set of a neural network $f_\theta\colon \mathbb{R}^d \times \mathbb{R}^3 \to \mathbb{R}$:
$$S^{(k)} = \{x \in \mathbb{R}^3 \colon f_\theta(z_k; x) = 0\}, \tag{2}$$
where $\theta$ are trainable parameters and $z_k \in \mathbb{R}^d$ are trainable embedding vectors, which encode the geometric properties of each shape. The proposed network has an auto-decoder architecture and is trained on a set of pairs consisting of points and corresponding SDF values sampled from a neighborhood of the surface and the ambient.

For smooth surfaces, the signed distance field $f$ is differentiable and satisfies the eikonal PDE $\|\nabla f(x)\| = 1$. Taking advantage of this property, Gropp et al. [1] improved the previously mentioned technique by introducing an implicit geometric regularization method, which makes the network $f_\theta$ produce a signed distance field by encouraging the spatial gradient of $f_\theta$ to have unit norm. This regularization allows for ground truth points to be sampled only from the surface.

## 2. Methods

Based on the works of Park et al. [5] and Gropp et al. [1], we develop a DL-based aortic shape generation model described together with the training dataset as follows.

**Model.** Following Park et al. [5] and Gropp et al. [1], we represent each of the $n$ shapes as the zero-level set of a neural signed distance field $f_\theta$. Our network is an 8-layer fully connected auto-decoder, with softplus activation functions. As illustrated in Figure 1, we add a skip-connection after the fourth layer, which improves stability and allows the network to utilize high-dimensional embedding vectors effectively (Rebain et. al. [6]).

The loss function consists of three terms. The first term is responsible for making the network $f_\theta$ vanish on points sampled from the surface and enforcing the consistency of normal vectors. The second term is the implicit geometric regularization, which makes $f_\theta$ produce a signed distance field by encouraging the spatial gradient of $f_\theta$ to have unit norm, and the last term is an $L^2$ regularization for the embedding vectors. Concretely, the loss has the expression:

$$\mathcal{L}(\theta, z_1, \ldots, z_n) = \frac{1}{n}\sum_{k=1}^{n} \mathcal{L}^{(k)}(\theta, z_k) \quad (3)$$

where $\mathcal{L}^{(k)}$ is the loss corresponding to the $k$-th shape, given by

$$\mathcal{L}^{(k)}(\theta, z_k) = \frac{1}{N}\sum_{i=1}^{N}\left(|f_\theta(z_k; x_i^{(k)})| + \|\nabla_x f_\theta(z_k; x_i^{(k)}) - n_i^{(k)}\|^2\right) \\ + \tau \mathbb{E}_{x \sim D}[(\|\nabla_x f_\theta(z_k; x)\| - 1)^2] + \lambda \|z_k\|, \quad k = \overline{1, n}. \quad (4)$$

The expectation in (4) is estimated by sampling points $x \in \mathbb{R}^3$ from a distribution $D$ which is the average between a sum of Gaussian distributions centered at the sampled points and a uniform distribution.

For training the network, we use the Adam optimizer for 5000 epochs with an initial learning rate of $10^{-3}$, which is decreased by a factor of 0.5 every 500 epochs. The embedding vectors have dimension $d = 256$ and are initialized from a normal distribution with mean 0 and standard deviation $10^{-2}$. The coefficients in (4) are chosen $\tau = 0.5$ and $\lambda = 10^{-4}$.

**Dataset.** The training dataset consists of $n = 97$ aortic root (non-watertight) meshes normalized to a unit ball. For each mesh, we sample $N = 500000$ surface points along with unit normal vectors $\left\{(x_i^{(k)}, n_i^{(k)}) \in \mathbb{R}^3 \times \mathbb{R}^3, i = \overline{1, N}\right\}, k = \overline{1, n}$.

**Shape reconstruction and generation.** For an embedding vector $z$, the corresponding mesh can be recovered by running Marching Cubes on the SDF values $f_\theta(z, x)$ produced by the network on a uniform $256^3$ grid. In addition, new shapes can be generated by taking $z$ to be a convex combination (weighted average) of the learned embedding vectors:

$$z = \alpha_1 z_1 + \alpha_2 z_2 + \cdots + \alpha_n z_n, \quad (5)$$

with $\alpha_1, \ldots, \alpha_n \geq 0$ and $\alpha_1 + \cdots + \alpha_n = 1$.

## 3. Results

**Shape reconstruction.** For verifying the fidelity of the learned representations, we reconstruct the meshes from the training set using the learned embedding vectors and evaluate the similarity using the Chamfer distance (Fan et al. [2]) computed on sets of points sampled from each mesh.

The distribution of Chamfer distances between the ground truth and reconstructed shapes is illustrated in Figure 2. We observe that almost all shapes can be accurately reconstructed, except for one outlier, which upon manual inspection revealed a reconstruction artifact near the outlet.

**Shape generation.** To evaluate the novelty of the generated shapes, we study the distribution of pairwise Chamfer distances (illustrated in Figure 3) on a set of 100 new shapes, generated by interpolating between 2, 4, and 8 embedding vectors. We observe that increasing the number of interpolated vectors decreases the variance of the distribution, because the novel shapes better resemble existing shapes in the dataset. When using minimum two vectors for interpolation, the distribution trades less variance, and the shapes generated are more diverse. An illustration of 3 generated samples is provided in Figure 4. Each shape is generated from a new embedding vector constructed by interpolating between two learned embedding vectors $z_1$ (corresponding to the shape in Figure 4a) and $z_2$, with $\alpha_1 = \alpha_2 = 0.5$.

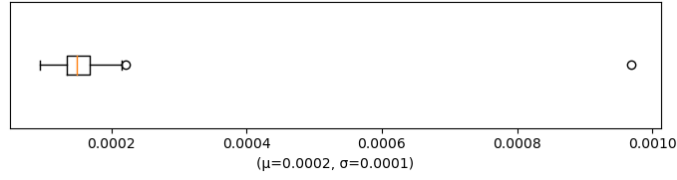

**Figure 2.** Reconstruction Chamfer distance distribution.

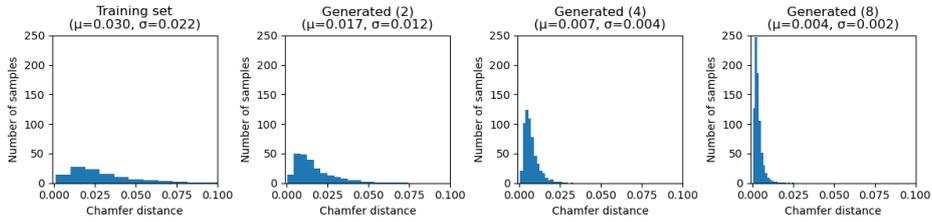

**Figure 3.** Pairwise Chamfer distance distribution.

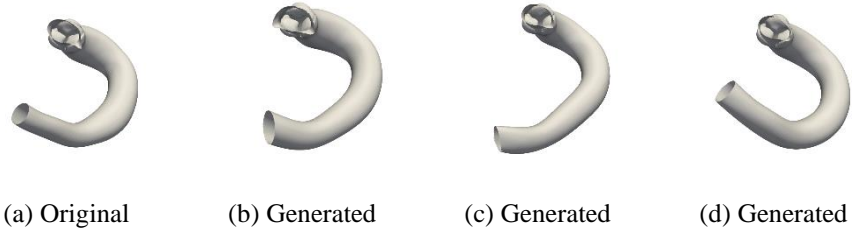

(a) Original  (b) Generated  (c) Generated  (d) Generated

**Figure 4.** Examples of shapes generated with the proposed method.

## 4. Discussion

Our initial experiments show that the shapes produced by this method exhibit less variance, and thus less diversity, compared to the original training data. This reduced variance results from the generative process, which relies on interpolating pre-trained embedding vectors. Adjusting the number of embedding vectors involved in the generation process can help manage the trade-off between variance and diversity. Our findings suggest that the number of samples chosen for interpolation significantly impacts the diversity of the generated cohorts.

Exploring innovative generative techniques represents a promising direction for future research. For instance, selecting vectors based on their variance relative to the initial training distribution could enhance the variance in synthetic datasets. Moreover, introducing noise during the generation process may further increase cohort diversity. Clinicians can then fine-tune these hyperparameters to customize the cohorts for specific in-silico trials. Additionally, incorporating anatomical measurements [7] may provide a more precise way to assess the variance of synthetic cohorts and could be used to generate shapes with specific anatomical features.

## 5. Conclusions

Our study demonstrates the potential of using DL-based methods to generate aortic shapes for cardiovascular in-silico trials. While our method effectively generates synthetic cohorts, it may exhibit limitation in variance and diversity due to its reliance on interpolating pre-trained embedding vectors. To address these drawbacks, future work could explore more advanced generative techniques such as generative adversarial networks or diffusion models and incorporate attention mechanisms for conditioning the neural field. Additionally, expanding the scope of our model to include or generate features such as calcifications or predictions of velocity and pressure could significantly enhance its accuracy and efficacy. These improvements would enable clinicians to tailor synthetic cohorts for specific in-silico trials, leading to more personalized outcomes.

## Acknowledgements

This project has received funding from the European Union's Horizon 2020 research and innovation program under grant agreement No 101017578.